\documentclass[a4paper,conference]{IEEEtran}
\usepackage[table]{xcolor}
\usepackage{tikz}
\usepackage{pgfplots}
\usepackage{pgfplotstable}
\usepackage{multicol}
\usepackage{amssymb}
\usepackage{diagbox}
\usepackage{gensymb}
\usepackage{hyperref}
\usepgfplotslibrary{colormaps}

\newcolumntype{C}{>{\centering\arraybackslash}p{6mm}}

\usepackage{amsmath}
\makeatletter
\def\resetMathstrut@{%
  \setbox\z@\hbox{%
    \mathchardef\@tempa\mathcode`\[\relax
    \def\@tempb##1"##2##3{\the\textfont"##3\char"}%
    \expandafter\@tempb\meaning\@tempa \relax
  }%
  \ht\Mathstrutbox@\ht\z@ \dp\Mathstrutbox@\dp\z@}
\makeatother
\begingroup
  \catcode`(\active \xdef({\left\string(}
  \catcode`)\active \xdef){\right\string)}
\endgroup
\mathcode`(="8000 \mathcode`)="8000

\definecolor{darkgreen}{rgb}{0,0.5,0}

\title{Augmentation Inside the Network}

\author{
\IEEEauthorblockN{Maciej Sypetkowski}
\IEEEauthorblockA{Faculty of Mathematics, Informatics\\ and Mechanics\\
University of Warsaw\\
Email: m.sypetkowski@student.uw.edu.pl}
\and
\IEEEauthorblockN{Jakub Jasiulewicz}
\IEEEauthorblockA{Faculty of Mathematics, Informatics\\ and Mechanics\\
University of Warsaw\\
Email: j.jasiulewicz@student.uw.edu.pl}
\and
\IEEEauthorblockN{Zbigniew Wojna}
\IEEEauthorblockA{University College London\\
Tensorflight Inc.\\
Email: zbigniewwojna@gmail.com}
}

\def\eg{\emph{e.g.}}
\def\ie{\emph{i.e.}}

\def\etal{\emph{et al.}}

\def\reals{\mathbb{R}}

\def\sensitivity{inside-impact}
\def\Sensitivity{Inside-impact}

\newcommand{\plotticktext}[1]{\rotatebox{90}{\small \textit{#1}}}
\def\plotopacity{0.4}
\def\tableheight{-1pt}
\def\tablesize{\small }

\pgfplotsset{tick label style={font=\small},label style={font=\small},legend style={font=\footnotesize}}

\begin{document}

\maketitle

\begin{abstract}
In this paper, we present \textit{augmentation inside the network},
a method that simulates data augmentation techniques
for computer vision problems on intermediate
features of a convolutional neural network. We perform these transformations, changing the data flow through the network, and sharing common computations when it is possible.
Our method allows us to obtain smoother speed-accuracy trade-off adjustment and achieves better results than using standard test-time augmentation (TTA) techniques.
Additionally, our approach can improve
model performance even further when coupled with test-time augmentation.
We validate our method on the ImageNet-2012 and CIFAR-100 datasets
for image classification.
We propose a modification that is 30\% faster than the flip test-time augmentation and achieves the same results for CIFAR-100.

\end{abstract}

\section{Introduction}
Since the first successful uses of deep convolutional networks for the task of image classification ~\cite{lenet,alexnet} these architectures have dominated the field, and their various iterations are to this day regarded as state of the art. 

Much work and focus of researchers have been put into innovating and improving on previous network architectures ~\cite{resnet, wide-resnet, resnext, inception, xception, senet, densenet}.

Some problems can be tackled in the architecture, in particular, convolutional neural networks were designed to be unaffected by translations, and
Harmonic Networks \cite{harmonic-networks} are an architecture which is locally equivariant to patch-wise translation and rotation.

One other improvement is the use of data augmentation, which is standard in modern approaches to image classification, yet has seen much slower improvement over techniques used already in 2012 ~\cite{alexnet}.

State of the art computer vision techniques take advantage of test-time augmentation when aiming for the highest possible score ~\cite{detectron2, vgg}, at the cost of additional computation time during inference. Our goal in this work is to simulate TTA as part of the neural architecture.

Traditional data augmentation techniques based on performing separate calculations on multiple transformed copies of the input image can be regarded as branching the network out at the very start, performing independent calculations on each variant, and combining the outputs afterward during TTA. 

\begin{figure*}
    \centering
    \includegraphics[height=110pt]{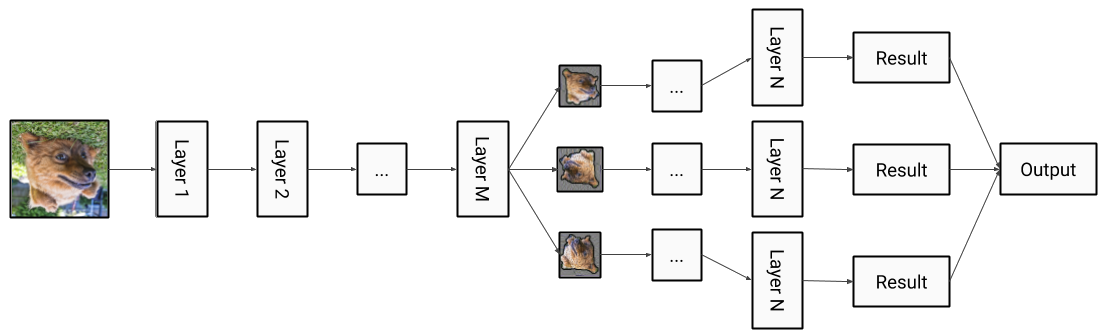}
    \caption{Augmentation inside the network.}
    \label{fig:augmentation_inside_the_network}
\end{figure*}

Our approach is a generalization of that idea, in the sense that the branching point is located inside the network -- the schema is shown in Figure \ref{fig:augmentation_inside_the_network}. Before that point, the input is run through the network in a standard way. When it reaches a branching point, multiple variants of the data are created that are the result of various augmentation-like transformations on the output from the previous layer. Then, each variant is run through the remaining part of the network independently. Just like with standard augmentation, all of those branches share their weights. 

The process can be applied either for training the network or during inference, effectively simulating both data augmentation and test-time augmentation. For inference, in particular, results from all the branches are combined to create a single output. For training, each output can be treated separately.

Our contributions include:
\begin{itemize}
\item Exploration of the problem of time-consuming test-time augmentation and introduction of a new method better suited to such scenarios. 
\item Analysis of the impact of the layer depth at which augmentations are performed on the overall performance with a trade-off between accuracy and computational time. 
\item A new architecture that achieves results superior to baselines. 
\item Combining our techniques with test-time augmentation to further boost accuracy.
\end{itemize}

\section{Related Work}

\textbf{Augmentation.}
Already in 1998, LeCun {\etal} used artificially distorted images to augment their training set for LeNet ~\cite{lenet}. 
Later, Krizhevsky {\etal}, while introducing the convolutional neural network AlexNet ~\cite{alexnet} for image recognition on the ImageNet dataset~\cite{imagenet}, used data augmentation techniques to help prevent overfitting and improve the accuracy. 
Specifically, they employed translations and horizontal reflections on patches extracted from a given input image to increase their effective training set by a factor of 2048 through the addition of interdependent derivatives of the original image. 
They also employed PCA on the RGB channels to help their network's predictions be invariant to changing color or intensity of illumination.

The augmentation techniques developed for image classification also found a use in other computer vision tasks, such as semantic segmentation ~\cite{fcn, unet}, as well as object detection ~\cite{rcnn, fastrcnn, fasterrcnn, yolo}.

Generative adversarial networks (GANs) are also commonly used in a manner similar to data augmentation, especially in medical tasks. GANs are useful in particular for denoising ~\cite{ctdenoise}, super-resolution ~\cite{superres} as well as image synthesis ~\cite{melanogan, bmaugment, foolradio, liveraugment}.

\textbf{Feature space augmentation.}
Feature space augmentation is a technique that performs data augmentation on intermediate representations. It consists of isolating low dimensional intermediate representations from a portion of the network, and using them to create a better model. 

DeVries and Taylor ~\cite{fsataylor} use an LSTM-based sequence autoencoder to arrive at a generic method, which proceeds as follows: first, each sample is encoded, then the resulting representation gets subjected either to added Gaussian noise, or gets combined with its $k$-nearest neighbors in the feature space in order to interpolate or extrapolate synthetic representations, and the resulting feature representations are finally used either as raw input for different learning tasks, or passed through the decoder to generate artificial input data. The authors find an improvement in accuracy on the MNIST and CIFAR-10 datasets.

Moreover, Wong {\etal} ~\cite{wong} found traditional input space augmentation to outperform feature space augmentation on MNIST not only for convolutional neural networks but also for convolutional support vector machines and convolutional extreme learning machines, with the improvement from using elastic deformation for augmenting the input space being surpassed only by adding real samples. 

In HoloGAN~\cite{hologan}, the idea of transforming the intermediate representation is used to generate new views of the same scene, by applying rigid transformations to existing 3D features. 

DropBlock~\cite{dropblock}, which can be regarded as a feature space augmentation variant of Cutout~\cite{cutout}, transforms feature maps by zeroing out a contiguous region of a given feature map. 

In this paper, we explore augmentation in the feature space by transforming intermediate features in a way similar to data augmentation.

\textbf{Speed-accuracy trade-off.}
There are many works that try to optimize speed and latency while still preserving good accuracy.
Mobile architectures ~\cite{mobilenet, mobilenetv2, shiftnet, shufflenet, shufflenetv2, squeezenet} were designed for devices that have lackluster computational power. Distillation methods ~\cite{distilling-the-knowledge} and pruning methods were also proposed as a means of making network calculations faster.
Many of the most accurate object detection architectures ~\cite{fastrcnn, fasterrcnn, maskrcnn, retinanet, cascadercnn} have difficulties running in realtime with low latency and high FPS. Because of that, faster but less accurate alternatives were created ~\cite{ssd, yolo}.
Many neural architecture search (NAS) methods were also proposed to automatically search for the best architecture given specific speed requirements ~\cite{mnasnet, mobilenetv3, fbnet, onceforall}.

Since the computational requirements of our techniques can be freely adjusted, our methods allow both for adjusting the speed and accuracy in computationally limited environments, as well as for improving accuracy in cases when it is more important than speed.

\section{Method}

\subsection{Method Formulation}

A convolutional network can be defined as a function:\footnote{for simplicity we assume batch size equals $1$}
\begin{align}
\mathcal{N}& : \reals^{1 \times H_0 \times W_0 \times C_0} \longrightarrow \reals^{1 \times H_k \times W_k \times C_k} \\
\mathcal{N}&(X) = (\mathcal{F}_k \circ \ldots \circ \mathcal{F}_1)(X) = \underset{i=1 \ldots k}{\bigcirc} \mathcal{F}_i (X)
\end{align}
where $\mathcal{F}_i$ is a function defining the $i$-th building block of the network and
$X \in \reals^{1 \times H_0 \times W_0 \times C_0}$ is an input image.
$H_i$ and $W_i$ are spatial dimensions and $C_i$ is the number of channels of the $i$-th feature map.
In particular, $H_0$, $W_0$, $C_0$ are dimensions of the input image.

We define a \emph{Branched Convolutional Network} (or simply a branched network) as 
\begin{align}
\mathcal{B}& : \reals^{1 \times H_0 \times W_0 \times C_0} \longrightarrow \reals^{R \times H_k \times W_k \times C_k} \\
\mathcal{B}&(X) = \underset{i=1 \ldots k}{\bigcirc} (\mathcal{A}_i \circ \mathcal{F}_i) (X)
\end{align}
where $R = \prod_{i=1 \ldots k} R_i$ and $\mathcal{A}_i$ is the $i$-th \emph{branching function}.
The branching function $\mathcal{A}_i : \reals^{B_i \times H_i \times W_i \times C_i} \longrightarrow \reals^{(R_i \cdot B_i) \times H_i \times W_i \times C_i}$
transforms the $i$-th feature map in the batch by independently augmenting it $R_i$ times using different methods.
In our experiments, most of $\mathcal{A}_i$ are identity functions.
Details about the set of transformations used are included in Section \ref{inside-augmentations}.

A Branched Convolutional Network is a generalization of a regular convolutional network that can return multiple variants of feature maps. Because of that, adjustments have to be made to the training and inference process.

\subsubsection{Training and Inference}
Let $X$ be an input image, and $(Y_1, \ldots, Y_R) = \mathcal{B}(X)$. Let $H$ be the head of the network ({\ie} the function that performs the final regression based on the feature map, {\eg} in classification problem it outputs the vector of probabilities for all classes), and $P_i = H(Y_i)$. 

\emph{Training}:
Let $\textit{criterion}$ be a criterion function ({\eg} cross entropy), and $T$ be the ground truth for that criterion function.
We use two main ways of calculating loss:
\begin{itemize}
    \item $\textit{loss} = \frac{\sum_{i=1 \ldots k}\textit{criterion}(P_i, T)}{k}$ -- loss is applied independently to every branch and then averaged. In experiments we will refer to it as \textit{none} reduction.
    \item $\textit{loss} = \textit{criterion}(\textit{reduction}(\{P_i \;|\; i = 1 \ldots k\}), T)$ -- We reduce outputs of branches using one of the methods described in Section \ref{reductions} ({\eg} \textit{max}), and then apply the \textit{criterion} to the reduced output and the ground truth.
\end{itemize}

\emph{Inference}:
Like in the second method for training we arrive at the output by reducing across all branches, {\ie}
$\textit{output} = \textit{reduction}(\{P_i \;|\; i = 1 \ldots k\})$.

\subsubsection{Reductions}
\label{reductions}
In most of our experiments we use \textit{max}, \textit{sum}, and \textit{geo} -- respectively taking maximum, arithmetic mean or geometric mean for each class independently from all branches. Note about naming: after applying the reduction, we normalize the resulting class probability vector. As a result, \textit{sum}ming before normalizing gives us the arithmetic mean.

Applying \textit{geo} reduction on the probability vector is the same as the arithmetic mean of the logits, if we calculate the probability vector as the softmax of the logits, since\footnote{We remove normalizing factor as we normalize the result after applying the reduction}

\begin{align*}
\sqrt[k]{\prod_{i=1\ldots k} \exp(P_{i,\textit{class}})} =
\sqrt[k]{\exp(\sum_{i=1\ldots k}P_{i,\textit{class}})} = \\
\exp(\frac{{\sum_{i=1\ldots k}P_{i,\textit{class}}}}{k})
\end{align*}

\subsubsection{Inside-augmentations}
\label{inside-augmentations}
The augmentations we perform are \textit{flip}, \textit{rotation}, and \textit{scale}.
For every transformation, we keep the dimensions of the feature map.
\textit{flip} consists of flipping feature maps horizontally. For zooming out, we pad the remaining part with zeros, and for zooming in, we crop the scaled feature maps. We use bilinear interpolation.
We focus mostly on \textit{flip} as we are able to obtain the best results from it.

\section{Experiments}

\subsection{CIFAR-100}
\label{cifar-experiments}
We conduct image classification experiments on the CIFAR-100~\cite{cifar} dataset.
We use the PreAct ResNet-110~\cite{preact-resnet} architecture.
All experiments are run for 180 epochs with 128 images per minibatch.
The learning rate starts at 0.1, and is divided by 10 after 82, 123 epoch and by 5 after 160 epoch.
The optimizer we use is SGD with $0.9$ momentum. Weight decay is set to $2 \cdot 10^{-4}$.
The slowdown was measured on an NVIDIA GTX 1080 Ti GPU.

\textit{vanilla} is the baseline configuration without our method.
\textit{vanilla-tta-red} is a baseline configuration with standard horizontal flip test-time augmentation with the reduction specified as \textit{red}.
\textit{flip-$n$-red} is a configuration using our method with the \textit{flip} inside-augmentation branching on each of the last $n$ feature maps before the last convolution (which results in $2^n$ final feature map variants) and using reduction \textit{red} during both training and inference. Adding \textit{-tta} means combining our method with standard test-time augmentation.

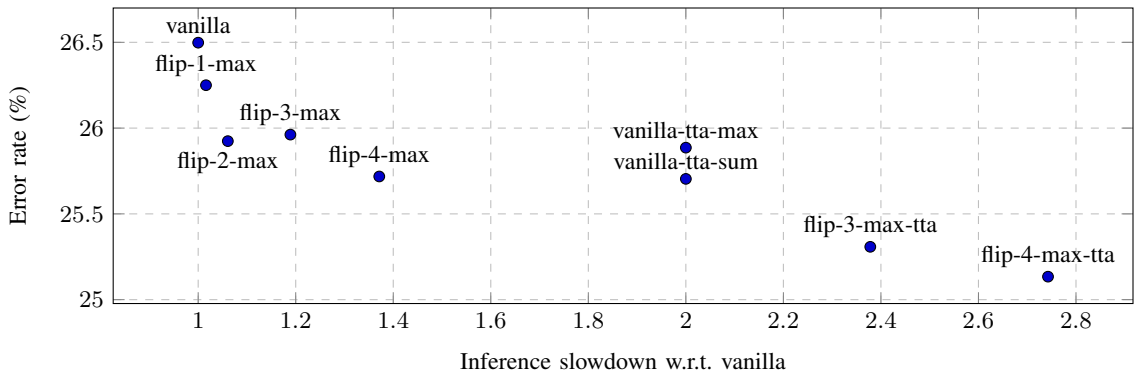
\begin{figure*}
\center
\begin{tikzpicture}
\begin{axis}[
    xlabel={Inference slowdown w.r.t. vanilla},
    ylabel={Error rate (\%)},
    ymajorgrids=true,
    xmajorgrids=true,
    grid style=dashed,
    nodes near coords,
    width=15cm, height=5.5cm,
    ymax=26.7,
]

\addplot+[only marks, point meta=explicit symbolic, color=black]
    coordinates {
(1,26.498)[\small vanilla]
(2,25.886)[\small vanilla-tta-max]
(2,25.704)[\small vanilla-tta-sum]
(1.01601164483261,26.25)[\small flip-1-max]

(1.06079027355623,25.924)

(1.18909710391823,25.962)[\small flip-3-max]
(2.37819420783646,25.308)[\small flip-3-max-tta]

(1.3713163064833,25.718)[\small flip-4-max]
(2.7426326129666,25.134)[\small flip-4-max-tta]

};

\node[] at (axis cs: 1.06079027355623,25.8) {\small flip-2-max};
    
\end{axis}
\end{tikzpicture}

\caption{
Results on CIFAR-100 dataset using PreAct ResNet-110.
Each point is the average over 5 runs.
Of particular note is \textit{flip-4-max}, as it obtains the same accuracy as vanilla with test-time augmentation, and is more than 30\% faster.
Moreover, combining our method with test-time augmentation improves results even further.
}
\label{fig:cifar100-results}
\end{figure*}

We summarize our results in Figure \ref{fig:cifar100-results}.

\subsection{ImageNet}
\label{imagenet-experiments}
Next, we do experiments on ImageNet-2012.
We use ResNet-18~\cite{resnet}.
We train for 105 epochs with a batch size of 256.
The learning rate starts at $0.1$ and gets divided by 10 at epochs 30, 60, 90, 100.
We use a weight decay of $10^{-4}$. For augmentation, we follow Krizhevsky {\etal} ~\cite{alexnet} and additionally use random brightness, contrast, and saturation with a coefficient of $0.4$.
The optimizer we use is SGD with $0.9$ momentum with Nesterov's accelerated gradient.

The naming scheme for configurations is the same as in the previous section. Additionally, we perform experiments where the reduction used in training and inference can differ. For instance \textit{max,sum} reduction means \textit{max} during training but \textit{sum} during inference.
\textit{flip-only2-red} branches only at the second to last feature map.

\begin{figure*}

\centering
\begin{tikzpicture}[scale=1]
\begin{axis}[
    ylabel={Top1 error rate (\%)},
    ymajorgrids=true,
    xmajorgrids=true,
    grid style=dashed,
    nodes near coords,
    width=15cm, height=5cm,
    xmin=0.9, xmax=2.15,
    ymin=28.7, ymax=29.78,
]

\addplot+[only marks, point meta=explicit symbolic, color=black]
    coordinates {
(1,29.632)[\small vanilla]
(2,28.844) 
(2,28.974)[\small vanilla-tta-sum]
(1.784,28.884)[\small flip-3-max,sum]
(1.784,29.152)[\small flip-3-none,geo]
(1.24581005586592,29.268) 
(1.24581005586592,29.288)[\small flip-only2-none,geo]

};

\node[] at (axis cs: 2,28.78) {\small vanilla-tta-geo};
\node[] at (axis cs: 1.24581005586592,29.15) {\small flip-only2-max,sum};
    
\end{axis}
\end{tikzpicture}

\centering
\begin{tikzpicture}[scale=1]
\begin{axis}[
    xlabel={Inference slowdown w.r.t. vanilla},
    ylabel={Top5 error rate (\%)},
    ymajorgrids=true,
    xmajorgrids=true,
    grid style=dashed,
    nodes near coords,
    width=15cm, height=5cm,
    xmin=0.9, xmax=2.15,
    ymin=9.85, ymax=10.48,
]

\addplot+[only marks, point meta=explicit symbolic, color=black]
coordinates {
(1,10.374)[\small vanilla]
(2,9.926) 
(2,9.966)[\small vanilla-tta-sum]
(1.784,9.922)[\small flip-3-max,sum]
(1.784,10.236)[\small flip-3-none,geo]
(1.24581005586592,10.218)[\small flip-only2-max,sum]
(1.24581005586592,10.318)[\small flip-only2-none,geo]

};

\node[] at (axis cs: 2,9.88) {\small vanilla-tta-geo};
    
\end{axis}
\end{tikzpicture}


\caption{
Results on ImageNet-2012 dataset using ResNet-18. Each experiment was run 2 times. We obtain results between \textit{vanilla} and \textit{vanilla-tta} in both accuracy and inference time.
}
\label{fig:resnet18-results}
\end{figure*}
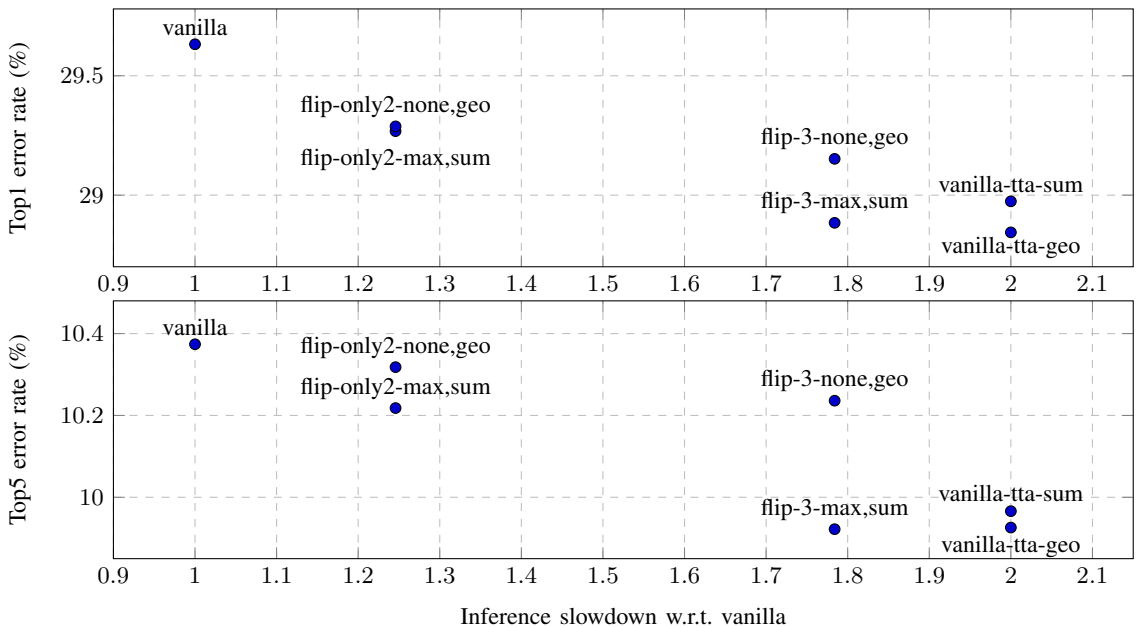

The results are summarized in Figure \ref{fig:resnet18-results}.

\subsection{\Sensitivity}
\label{sensitivity}
To better understand the intermediate features of the network,
we measure a quality which we call \textit{\sensitivity}.
For every potential spot where an inside-augmentation could be performed in the network,
we measure how much transforming the intermediate features in that spot affects accuracy.

We focus on two separate types of {\sensitivity} measurements.
\textit{Inference \sensitivity} tells us how transforming the features only in a given spot inside the network affects accuracy,
assuming that the network was trained without any transformations.
We do not branch inside the network. We only transform the feature map once at a given spot.

\textit{Training \sensitivity} tells us what accuracy we obtain when augmenting features in a given spot both during training and during inference.
We branch into two separate batches at that point, augmenting both of them, and reducing them during inference.
All experiments were trained in the way described in Sections \ref{cifar-experiments} and \ref{imagenet-experiments}.

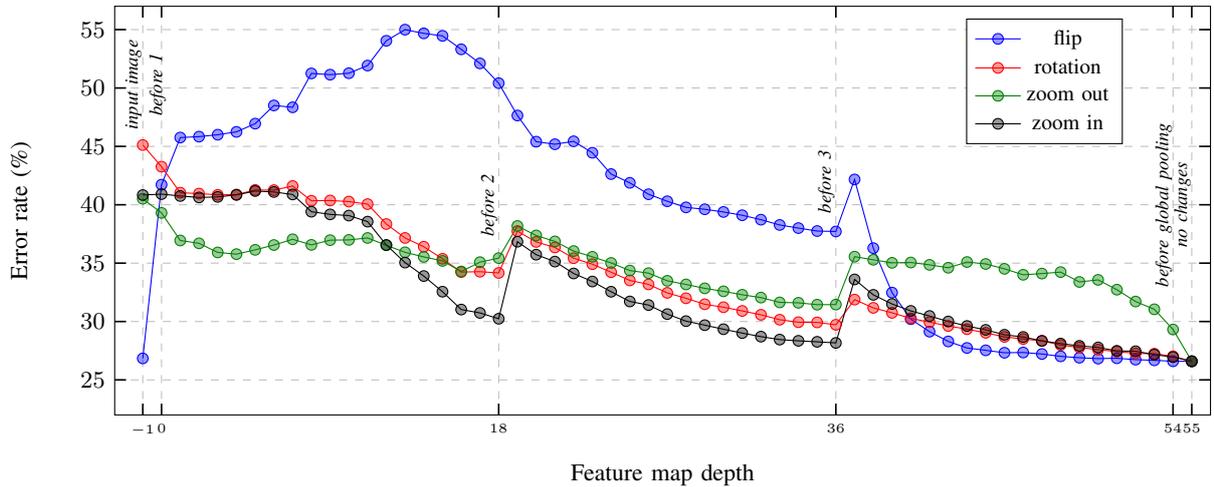
\begin{figure*}

\hspace*{-18pt}
\centering
\begin{tikzpicture}

\begin{axis}[
    xlabel=Feature map depth, ylabel=Error rate (\%),
    xticklabel style={font=\tiny},
    width=16cm, height=7cm,
    xmin=-2.5, xmax=56, ymin=22, ymax=57,
    grid=major,
    grid style=dashed,
    xtick={-1,0,18,36,54,55},
    ytick={25,30,35,40,45,50,55},
    every tick/.style={black,semithick},
    legend style={at={(0.92,0.97)}, anchor=north east}]
]
    \addplot[color=blue, mark=*, fill opacity=\plotopacity] coordinates {
    (-1,26.842) (0,41.72) (1,45.754) (2,45.834) (3,45.998) (4,46.242) (5,46.948) (6,48.506) (7,48.346) (8,51.248) (9,51.14) (10,51.266) (11,51.922) (12,54.042) (13,54.996) (14,54.678) (15,54.464) (16,53.312) (17,52.108) (18,50.412) (19,47.644) (20,45.402) (21,45.182) (22,45.434) (23,44.452) (24,42.628) (25,41.878) (26,40.9) (27,40.298) (28,39.776) (29,39.612) (30,39.382) (31,39.1) (32,38.722) (33,38.27) (34,38.002) (35,37.748) (36,37.712) (37,42.166) (38,36.284) (39,32.476) (40,30.202) (41,29.132) (42,28.294) (43,27.736) (44,27.532) (45,27.324) (46,27.346) (47,27.218) (48,27.016) (49,26.9) (50,26.822) (51,26.854) (52,26.734) (53,26.676) (54,26.594) (55,26.594)
    };
    \addplot[color=red, mark=*, fill opacity=\plotopacity] coordinates {
    (-1,45.116) (0,43.262) (1,41.032) (2,40.956) (3,40.848) (4,40.872) (5,41.266) (6,41.25) (7,41.604) (8,40.34) (9,40.362) (10,40.274) (11,40.054) (12,38.348) (13,37.156) (14,36.406) (15,35.372) (16,34.236) (17,34.266) (18,34.16) (19,37.742) (20,36.818) (21,36.344) (22,35.424) (23,34.91) (24,34.224) (25,33.52) (26,33.178) (27,32.452) (28,31.99) (29,31.476) (30,31.224) (31,30.906) (32,30.574) (33,30.154) (34,29.94) (35,29.924) (36,29.718) (37,31.876) (38,31.168) (39,30.744) (40,30.272) (41,29.948) (42,29.618) (43,29.342) (44,29.034) (45,28.684) (46,28.506) (47,28.34) (48,27.988) (49,27.776) (50,27.59) (51,27.43) (52,27.272) (53,27.24) (54,27.022) (55,26.594)
    };
    \addplot[color=darkgreen, mark=*, fill opacity=\plotopacity] coordinates {
    (-1,40.51) (0,39.302) (1,36.938) (2,36.696) (3,35.916) (4,35.77) (5,36.138) (6,36.55) (7,37.042) (8,36.562) (9,36.968) (10,36.988) (11,37.156) (12,36.528) (13,35.928) (14,35.532) (15,35.192) (16,34.284) (17,35.084) (18,35.43) (19,38.18) (20,37.368) (21,36.86) (22,36.034) (23,35.542) (24,35.018) (25,34.364) (26,34.142) (27,33.486) (28,33.174) (29,32.84) (30,32.586) (31,32.29) (32,32.05) (33,31.636) (34,31.602) (35,31.442) (36,31.45) (37,35.554) (38,35.284) (39,35.026) (40,35.054) (41,34.854) (42,34.604) (43,35.094) (44,34.928) (45,34.512) (46,34.002) (47,34.106) (48,34.232) (49,33.388) (50,33.566) (51,32.726) (52,31.712) (53,31.046) (54,29.318) (55,26.594)
    };
    \addplot[color=black, mark=*, fill opacity=\plotopacity] coordinates {
    (-1,40.832) (0,40.912) (1,40.75) (2,40.626) (3,40.672) (4,40.842) (5,41.176) (6,41.092) (7,40.882) (8,39.406) (9,39.176) (10,39.074) (11,38.554) (12,36.556) (13,35.048) (14,33.894) (15,32.554) (16,31.024) (17,30.736) (18,30.236) (19,36.834) (20,35.724) (21,35.132) (22,34.114) (23,33.426) (24,32.55) (25,31.718) (26,31.41) (27,30.636) (28,30.032) (29,29.686) (30,29.34) (31,29.008) (32,28.706) (33,28.452) (34,28.346) (35,28.262) (36,28.168) (37,33.604) (38,32.282) (39,31.502) (40,30.908) (41,30.472) (42,30) (43,29.616) (44,29.284) (45,28.866) (46,28.678) (47,28.344) (48,28.11) (49,27.918) (50,27.778) (51,27.488) (52,27.454) (53,27.166) (54,26.954) (55,26.594)
    };
    
    \legend{flip, rotation, zoom out, zoom in}
    
    \node[] at (axis cs: -1.5,50.3) {\plotticktext{\scriptsize input image}};
    \node[] at (axis cs: -0.2,50.3) {\plotticktext{\scriptsize before 1}};
    \node[] at (axis cs: 17.5,40) {\plotticktext{\scriptsize before 2}};
    \node[] at (axis cs: 35.5,42) {\plotticktext{\scriptsize before 3}};
    \node[] at (axis cs: 53.5,40) {\plotticktext{\scriptsize before global pooling}};
    \node[] at (axis cs: 54.5,40) {\plotticktext{\scriptsize no changes}};

\end{axis}
\end{tikzpicture}

\caption{
    Inference {\sensitivity} of an unmodified PreAct ResNet-110 on CIFAR-100.
    \textit{before $n$} marks the feature map before the $n$-th pooling layer of the network (specific for the architecture used).
    Inside-augmentations at the beginning of the network have much bigger negative impact on the accuracy.
}
\label{fig:rn110-sensitivity}

\end{figure*}

In inference {\sensitivity} experiments, we use rotation by $15\degree$ and 10\% scale for zooming in and out. We summarize our experiments in Figures \ref{fig:rn110-sensitivity} and \ref{fig:rn18-sensitivity}.
We observe the convolutional filters in the first layers are very flip dependent, {\ie} they correspond to the specific side of the object.
The deeper the intermediate representation is, the more invariant it gets to changes.
The filters in the first layers at the given resolution ({\ie} after pooling layers) are much more sensitive to changes.

In training {\sensitivity} experiments, 
for \textit{rotation} during the training, the feature map is rotated by a uniformly random angle in the range $[-20\degree,20\degree]$ independently in both branches, and during the inference, one branch is not changed and the second is rotated by $15\degree$.
Similarly, for \textit{scale}, during the training, the feature map is scaled by a uniformly random scale in the range $[0.8, 1.25]$ independently in both branches, and during inference one branch is not changed and the second one is scaled non-deterministically by a scale from the same range as in training.

\begin{figure*}

\centering
\begin{tikzpicture}

\begin{axis}[
    xlabel=Feature map depth, ylabel=Error rate (\%),
    width=15cm, height=7cm,
    xmin=-2.2, xmax=29, ymin=25, ymax=29,
    grid=major,
    grid style=dashed,
    xtick={-1,0,9,18,27,28},
    every tick/.style={black,semithick},
    legend style={at={(0.905,0.03)}, anchor=south east}]
]
    \addplot[color=blue, mark=*, fill opacity=\plotopacity] coordinates {
    (-1,27.528) (0,27.444) (1,27.368) (2,27.576) (3,27.414) (4,27.424) (5,27.512) (6,27.184) (7,27.378) (8,27.268) (9,27.302) (10,27.44) (11,27.532) (12,27.464) (13,27.61) (14,27.458) (15,27.566) (16,27.608) (17,27.662) (18,27.604) (19,27.798) (20,27.588) (21,27.85) (22,27.806) (23,27.856) (24,27.902) (25,28.076) (26,27.99) (27,28.052) (28,28.012)
    };
    \addplot[color=red, mark=*, fill opacity=\plotopacity] coordinates {
    (-1,26.706) (0,26.906) (1,26.836) (2,26.7) (3,26.916) (4,26.714) (5,26.77) (6,26.666) (7,26.788) (8,26.736) (9,26.714) (10,26.99) (11,27.202) (12,26.956) (13,26.808) (14,27.078) (15,27.328) (16,26.934) (17,26.964) (18,27.132) (19,27.236) (20,27.604) (21,27.77) (22,27.708) (23,27.652) (24,27.726) (25,27.972) (26,27.99) (27,27.86) (28,28.012)
    };
    \addplot[color=darkgreen, mark=*, fill opacity=\plotopacity] coordinates {
    (-1,25.666) (0,25.914) (1,26.246) (2,26.366) (3,26.27) (4,26.148) (5,26.146) (6,26.192) (7,26.39) (8,26.602) (9,26.576) (10,26.648) (11,26.906) (12,26.478) (13,26.922) (14,27.118) (15,27.184) (16,27.254) (17,27.22) (18,27.228) (19,27.558) (20,28.04) (21,28.074) (22,28.18) (23,27.982) (24,27.878) (25,27.93) (26,27.902) (27,28.126) (28,28.012)
    };
    \addplot[color=black, mark=*, fill opacity=\plotopacity] coordinates {
    (-1,28.474) (0,28.558) (1,28.548) (2,28.402) (3,28.252) (4,28.348) (5,28.266) (6,28.45) (7,28.416) (8,28.368) (9,28.372) (10,28.45) (11,28.534) (12,28.486) (13,28.432) (14,28.432) (15,28.186) (16,28.544) (17,28.244) (18,28.26) (19,28.336) (20,28.004) (21,27.928) (22,27.816) (23,27.626) (24,27.626) (25,27.63) (26,27.76) (27,27.992) (28,28.086)
    };

    \legend{{rotation-none,geo}, {scale-none,geo}, {flip-none,geo}, {flip-max,max}}
    
    \node[] at (axis cs: -1.6,27) {\plotticktext{input image}};
    \node[] at (axis cs: -0.3,26.37) {\plotticktext{before 1}};
    \node[] at (axis cs: 8.7,26.1) {\plotticktext{before 2}};
    \node[] at (axis cs: 17.7,26.5) {\plotticktext{before 3}};
    \node[] at (axis cs: 26.7,26.7) {\plotticktext{before global pooling}};
    \node[] at (axis cs: 27.7,27) {\plotticktext{no changes}};

\end{axis}
\end{tikzpicture}

    \caption{
        Training {\sensitivity} on the CIFAR-100 dataset using PreAct ResNet-56.
        Early in the network, configurations with \textit{none,geo} reduction perform better than near the end. When using \textit{flip-max,max} the situation is reversed as it performs a little better near the end of the network, slightly outperforming other configurations. 
    }
    \label{fig:rn56-sensitivity}
\end{figure*}
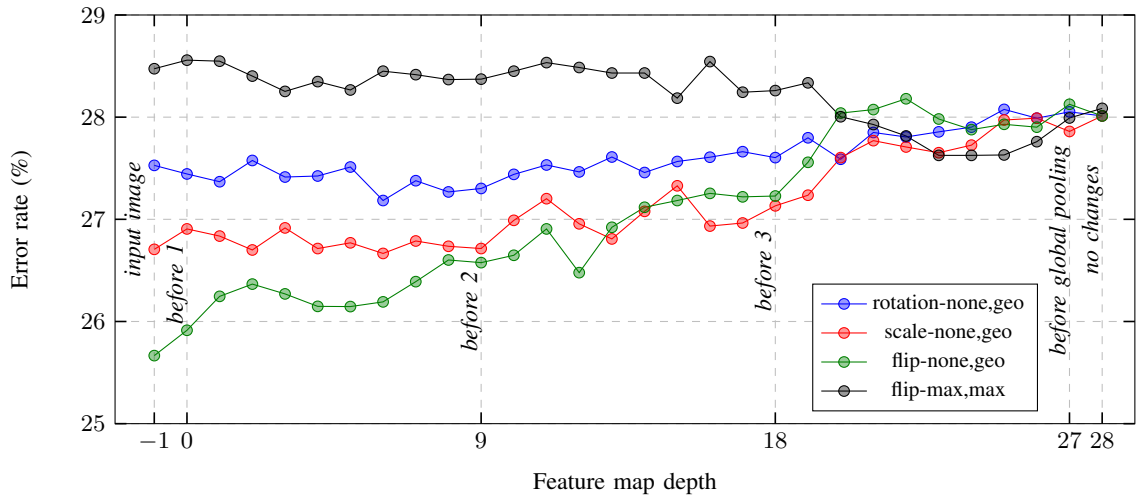

Figure \ref{fig:rn56-sensitivity} shows the results.
Flip is the most beneficial augmentation. All three augmentations improve when a shallower feature map is tested, together with increasing computational overhead, which allows for a very smooth speed-accuracy trade-off adjustment.


\begin{figure*}

\centering
\begin{tikzpicture}

\begin{axis}[
    xlabel=Feature map depth, ylabel=Error rate (\%),
    width=13cm, height=6.5cm,
    xmin=-2, xmax=11, ymin=0, ymax=100,
    grid=major,
    grid style=dashed,
    xtick={-1,1,3,5,7,9,10},
    every tick/.style={black,semithick},
]
    \addplot[color=blue, mark=*, fill opacity=\plotopacity] coordinates {
    (-1,29.625) (0,78.673) (1,78.689) (2,87.933) (3,86.019) (4,69.237) (5,48.807) (6,38.613) (7,32.463) (8,31.906) (9,29.632) (10,29.632)
    };
    \addplot[color=red, mark=*, fill opacity=\plotopacity] coordinates {
    (-1,10.413) (0,59.537) (1,59.735) (2,74.345) (3,70.414) (4,45.789) (5,24.259) (6,16.324) (7,12.183) (8,11.821) (9,10.406) (10,10.406)
    };
    
    \legend{top1 error, top5 error}
    
    \node[] at (axis cs: -1.25,50) {\plotticktext{input image}};
    \node[] at (axis cs: 0.75,50) {\plotticktext{before 1}};
    \node[] at (axis cs: 2.75,50) {\plotticktext{before 2}};
    \node[] at (axis cs: 4.75,50) {\plotticktext{before 3}};
    \node[] at (axis cs: 6.75,50) {\plotticktext{before 4}};
    \node[] at (axis cs: 8.75,50) {\plotticktext{before global pooling}};
    \node[] at (axis cs: 9.75,50) {\plotticktext{no changes}};

\end{axis}
\end{tikzpicture}

\caption{
    Inference {\sensitivity} for \textit{flip} using unmodified ResNet-18 on ImageNet.
    The plot looks consistent with \textit{flip} experiments presented in Figure \ref{fig:rn110-sensitivity} but is smoother and more contrasting in terms of accuracy.
    The last two points give the same error rate, since global pooling is invariant to flips.
}
\label{fig:rn18-sensitivity}
\end{figure*}
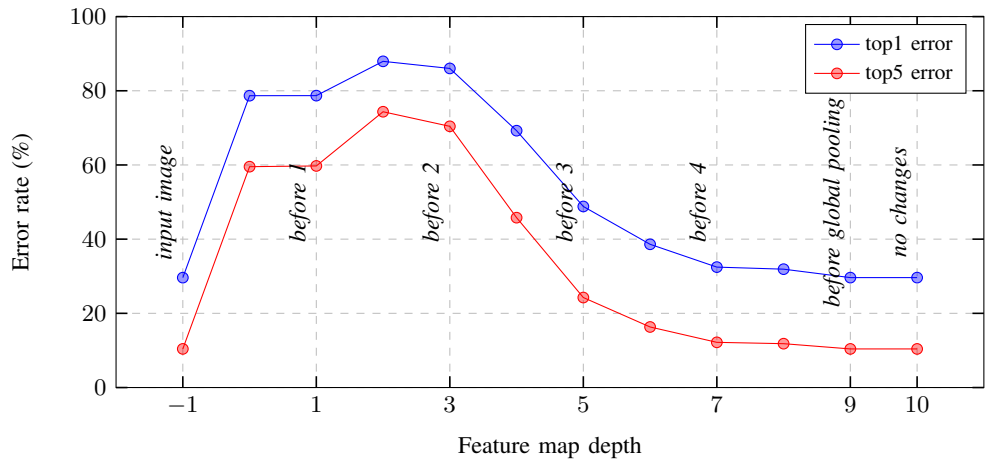

\subsection{Various Reduction Methods}
To compare how different reduction methods perform, we ran a series of experiments assuming different reduction methods independently for training and inference and inspecting the error rate.
We summarize our findings on CIFAR-100 in Figure \ref{fig:cifar100-reduce-matrix}, and for ImageNet in Figure \ref{fig:rn18-inference-reductions}. 
When training with \textit{max}, for inference \textit{sum} appears to be a safe default.
When training with \textit{none}, the best choice is \textit{geo}.
The results from the \ref{sensitivity} Section show that \textit{max} training reduction performs better in the later layers but does not perform well at the beginning and in the middle of the network where \textit{none} should be used instead.


\begin{figure}
     \centering
     \definecolor{custom}{RGB}{0,170,0}
     \begin{tabular}{rCCCC}

\diagbox[width=6em]{\scriptsize Training}{\scriptsize Inference}&\tablesize vanilla&\tablesize max&\tablesize sum&\tablesize geo\\[\tableheight] \tablesize vanilla&\tablesize 26.61\cellcolor{custom!26}&\tablesize 26.64\cellcolor{custom!24}&\tablesize 26.62\cellcolor{custom!25}&\tablesize 26.69\cellcolor{custom!21}\\[\tableheight] \tablesize none&\tablesize 26.40\cellcolor{custom!41}&\tablesize 26.39\cellcolor{custom!42}&\tablesize 26.40\cellcolor{custom!42}&\tablesize 26.40\cellcolor{custom!42}\\[\tableheight] \tablesize max&\tablesize 28.27\cellcolor{custom!0}&\tablesize 25.88\cellcolor{custom!99}&\tablesize 25.93\cellcolor{custom!91}&\tablesize 26.95\cellcolor{custom!9}\\[\tableheight] \tablesize sum&\tablesize 26.44\cellcolor{custom!39}&\tablesize 26.40\cellcolor{custom!42}&\tablesize 26.41\cellcolor{custom!41}&\tablesize 26.41\cellcolor{custom!41}\\[\tableheight] \tablesize geo&\tablesize 26.55\cellcolor{custom!30}&\tablesize 26.53\cellcolor{custom!32}&\tablesize 26.55\cellcolor{custom!30}&\tablesize 26.55\cellcolor{custom!30}\\[\tableheight]

    \end{tabular}
    \bigskip
    
    \caption{
        Mean error rate on CIFAR-100 using PreAct ResNet-110 with
        \emph{flip-3} augmentation and various ensembling functions for training and inference.
        Networks trained with \textit{none}, \textit{sum} or \textit{geo} perform very similarly for different inference reductions.
        Training with \textit{max} is different and leads to the best results with \textit{max} or \textit{sum} during inference.
    }
    \label{fig:cifar100-reduce-matrix}
\end{figure}

\begin{figure}
     \centering
     \definecolor{custom}{RGB}{0,170,0}
     \begin{tabular}{rCCCC}

\diagbox[width=6em]{\scriptsize Training}{\scriptsize Inference}&\tablesize vanilla&\tablesize max&\tablesize sum&\tablesize geo\\[\tableheight] \tablesize vanilla-tta&\tablesize 29.63\cellcolor{custom!0}&\tablesize 29.26\cellcolor{custom!10}&\tablesize 28.97\cellcolor{custom!56}&\tablesize 28.84\cellcolor{custom!100}\\[\tableheight] \tablesize flip-3-max&\tablesize 37.86\cellcolor{custom!0}&\tablesize 29.23\cellcolor{custom!17}&\tablesize 28.88\cellcolor{custom!100}&\tablesize 32.44\cellcolor{custom!0}\\[\tableheight] \tablesize flip-only2-max&\tablesize 31.16\cellcolor{custom!0}&\tablesize 29.49\cellcolor{custom!34}&\tablesize 29.27\cellcolor{custom!100}&\tablesize 29.51\cellcolor{custom!31}\\[\tableheight] \tablesize flip-3-none&\tablesize 29.46\cellcolor{custom!21}&\tablesize 29.29\cellcolor{custom!54}&\tablesize 29.18\cellcolor{custom!90}&\tablesize 29.15\cellcolor{custom!100}\\[\tableheight] \tablesize flip-only2-none&\tablesize 29.61\cellcolor{custom!19}&\tablesize 29.48\cellcolor{custom!42}&\tablesize 29.36\cellcolor{custom!72}&\tablesize 29.29\cellcolor{custom!100}\\[\tableheight]

    \end{tabular}
    \bigskip
    
    \caption{
        Mean top1 error rate on ImageNet using ResNet-18 with
        different augmentations and various reductions during inference.
        Colors in each row are scaled independently to compare different inference reductions across varied training configurations and not the configurations themselves.
        When training with \textit{none} reduction, both \textit{geo} and \textit{sum} perform well, while when training with \textit{max} reduction, \textit{sum} is the best for inference and is also noticeably better than the others.
    }
    \label{fig:rn18-inference-reductions}
\end{figure}

\section{Conclusion}

We propose a feature space augmentation method that applies transformations similar to data augmentation for computer vision
to intermediate representations inside the convolutional neural network.
We obtain different speed-accuracy trade-off adjustments depending on the number of performed transformations and their location in the network.
Some configurations lead to similar results as when using test-time augmentation but with faster inference.
Moreover, our method and standard test-time augmentation are not mutually exclusive and can be used together to produce even better results.

We study the {\sensitivity} of the transformations applied to intermediate features in different parts of the network.
The network is more sensitive at the beginning, especially to \textit{flip}s, and becomes progressively less sensitive closer to the end.
Similarly, applying an inside-augmentation at the beginning of the network may lead to a bigger improvement, but also to more significant computational overhead. 
The results from these experiments can also be a tool to gain a better understanding of the intermediate representations learned by the network.

We also study different reduction functions for both training and inference.
We find that a safe default is to train with \textit{max} and infer with \textit{sum} when deeper features are being transformed, or to train with \textit{none} and infer with \textit{geo} otherwise.

\section{Acknowledgements}
This work is an extension of research conducted as part of our Bachelor thesis at the University of Warsaw, which was coauthored by our colleagues Eryk Kijewski and Uladzislau Sobal, with help from Krzysztof Ciebiera.

\newpage



\end{document}